\documentclass[10pt,twocolumn,letterpaper]{article}

\usepackage[pagenumbers]{cvpr} 
\usepackage[table]{xcolor}
\usepackage[accsupp]{axessibility}

\definecolor{cvprblue}{rgb}{0.21,0.49,0.74}
\usepackage[pagebackref,breaklinks,colorlinks,allcolors=cvprblue]{hyperref}

\def\paperID{11563} 
\def\confName{CVPR}
\def\confYear{2026}

\title{PaQ-DETR: Learning Pattern and Quality-Aware Dynamic Queries for Object Detection}

\author{
Zhengjian~Kang$^{1}$\footnotemark[1],
Jun~Zhuang$^{2}$,
Kangtong~Mo$^{3}$,
Qi~Chen$^{4}$,
Rui~Liu$^{5}$,
Ye~Zhang$^{6}$\footnotemark[1]\kern0.33em\footnotemark[2]\\
$^1$\normalsize{New York University} \quad $^2$Boise State University
\quad $^3$\normalsize{University of Illinois at Urbana-Champaign}\\
$^4$\normalsize{University of California, Irvine} \quad $^5$\normalsize{Illinois Institute of Technology} \quad $^6$\normalsize{University of Pittsburgh}\\
}

\begin{document}
\maketitle
\footnotetext[1]{These authors contributed equally to this work.}
\footnotetext[2]{Corresponding author. Email: yez12@pitt.edu}
\begin{abstract}
Detection Transformer (DETR) has redefined object detection by casting it as a set prediction task within an end-to-end framework. Despite its elegance, DETR and its variants still rely on fixed learnable queries and suffer from severe query utilization imbalance, which limits adaptability and leaves the model capacity underused. We propose PaQ-DETR (Pattern and Quality-Aware DETR), a unified framework that enhances both query adaptivity and supervision balance. It learns a compact set of shared latent patterns capturing global semantics and dynamically generates image-specific queries through content-conditioned weighting. In parallel, a quality-aware one-to-many assignment strategy adaptively selects positive samples based on localization--classification consistency, enriching supervision and promoting balanced query optimization. Experiments on COCO, CityScapes, and other benchmarks show consistent gains of 1.5\%--4.2\% mAP across DETR backbones, including ResNet and Swin-Transformer. Beyond accuracy improvement, our method provides interpretable insights into how dynamic patterns cluster semantically across object categories.
\end{abstract}    
\section{Introduction}
\label{sec:intro}

\begin{figure}[t]
\centering
\includegraphics[width=\linewidth]{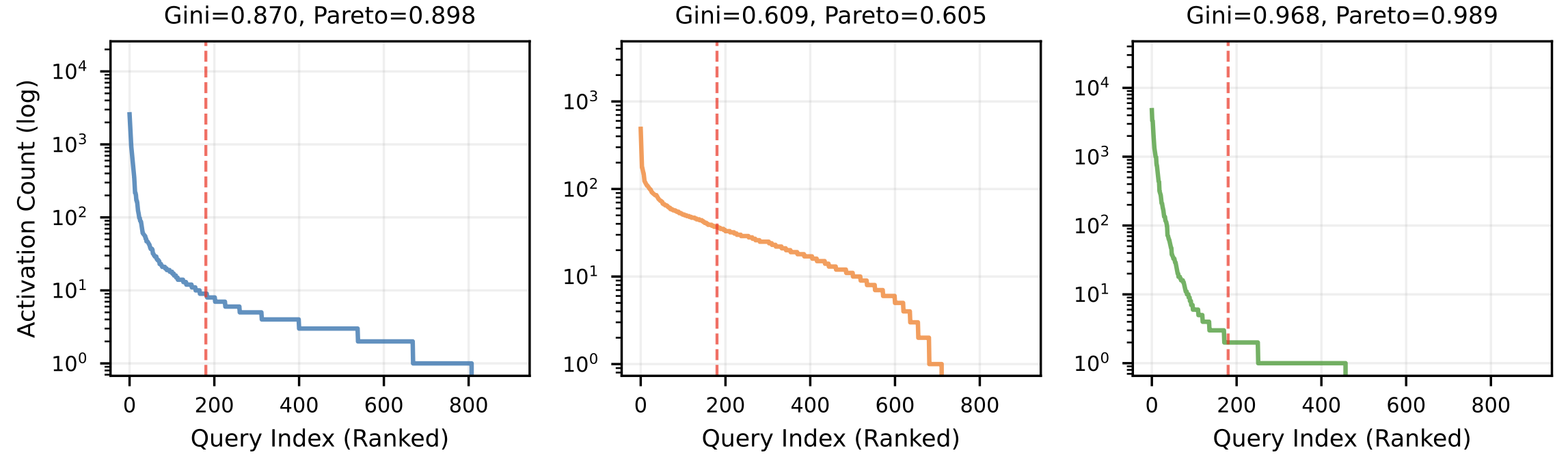}
\\[-3pt]
\makebox[0.33\linewidth][c]{\scriptsize\textit{(a) Deformable-DETR}}%
\makebox[0.33\linewidth][c]{\scriptsize\textit{(b) DN-DETR}}%
\makebox[0.33\linewidth][c]{\scriptsize\textit{(c) DINO}}
\\[2pt]
\includegraphics[width=\linewidth]{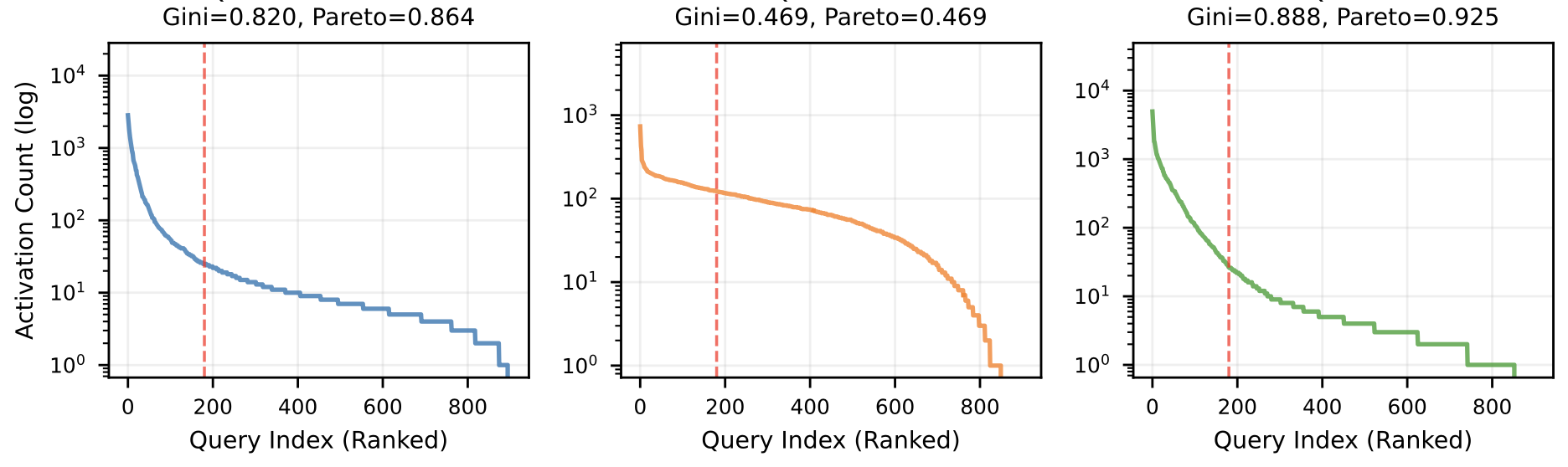}
\\[-3pt]
\makebox[0.33\linewidth][c]{\scriptsize\textit{(a) PaQ-Deformable-DETR}}%
\makebox[0.33\linewidth][c]{\scriptsize\textit{(b) PaQ-DN-DETR}}%
\makebox[0.33\linewidth][c]{\scriptsize\textit{(c) PaQ-DINO}}

\caption{
Query activation distributions of baseline DETRs vs. PaQ-DETR. 
Standard DETRs show highly skewed, long-tailed activations, while PaQ-DETR reduces imbalance and lowers Gini coefficients.
}
\vspace{-2em}
\label{fig:query_activation}
\end{figure}

Object detection~\cite{zou2023object}, widely used in autonomous driving~\cite{wang2025uniocc} and event-based perception~\cite{zhang2025adaptive}, aims to localize and classify objects within an image.
Traditional CNN-based detectors~\cite{Girshick2013RichFH,ren2016faster,Redmon2015YouOL} rely on dense anchors and heuristic post-processing (e.g., NMS), which hinder fully end-to-end learning. 
Transformers have redefined this paradigm: DETR~\cite{detr} formulates detection as a set prediction task trained with one-to-one Hungarian matching. 
Despite progress from follow-up variants~\cite{deformable,liu2022dab,li2022dn,dino,zong2023codetr,nan2024mi,kang2025lp}, two main challenges remain. 
First, content-dependent object queries improve adaptivity~\cite{deformable,lv2024rt,cui2023learning} but often exhibit unstable semantics across scenes~\cite{dino}. 
Second, the one-to-one matching scheme leads to highly sparse supervision: only a few queries consistently receive strong gradients, leaving many weakly optimized or under-used, which limits model capacity.

Prior works primarily pursue two directions for object queries:  
(1) static queries that maintain semantic stability but lack adaptability, and  
(2) content-dependent dynamic queries that increase flexibility but introduce instability.
However, these directions only partially address the underlying optimization issue.  
Our analysis shows that both challenges arise from a structural \emph{query activation imbalance} caused by DETR’s one-to-one matching, where a few ``winning'' queries receive most of the gradients while the majority remain weakly optimized.  
This perspective links query representation and supervision as two sides of the same problem, 
suggesting that mitigating imbalance requires jointly improving how queries are parameterized 
and how supervision signals are distributed.

To address this issue, we propose PaQ-DETR, which integrates two complementary components.  
First, we introduce a pattern-based dynamic query formulation that learns a compact set of shared patterns and composes image-specific queries via encoder-conditioned weighting. 
Unlike prior dynamic-query approaches (\eg, DDQ-DETR~\cite{zhang2023dense}, Dy-DETR~\cite{cui2023learning}, EASE-DETR~\cite{gao2024ease}), which improve adaptivity without explicitly addressing optimization imbalance, our formulation enables gradient sharing through a common representational basis, promoting more balanced and semantically coherent query evolution.  
Second, we introduce a quality-aware adaptive assignment strategy that selects positive samples according to localization--classification consistency, enriching supervision without auxiliary decoders or additional inference cost.

To validate the imbalance phenomenon, we analyze query activations across several DETR variants: Deformable-DETR~\cite{deformable}, DN-DETR~\cite{li2022dn}, and DINO~\cite{dino}. 
As shown in Fig.~\ref{fig:query_activation}, the activation distribution exhibits a substantial long-tail pattern, and the Gini coefficient reaches 0.97 in DINO, demonstrating severe inequality in query utilization and confirming the structural origin of imbalance.

Building on these insights, PaQ-DETR jointly improves query adaptivity and supervision balance, forming a unified optimization framework that enhances DETR training dynamics while keeping the architecture lightweight.

Our contributions are threefold:
\begin{itemize}[leftmargin=10pt,topsep=0pt]
    \item We empirically reveal and quantify severe \textbf{query activation imbalance} in DETR models 
    and trace its origin to the one-to-one Hungarian matching mechanism.
    \item We propose a \textbf{pattern-based dynamic query generation} mechanism 
    that unifies shared semantics and image-conditioned adaptivity, alleviating imbalance.
    \item We introduce a \textbf{quality-aware adaptive assignment} strategy 
    that balances supervision strength and further stabilizes optimization.
\end{itemize}

\section{Related Work}
\label{sec:related_work}

\begin{figure*}[t]
\centering
\includegraphics[width=1\textwidth]{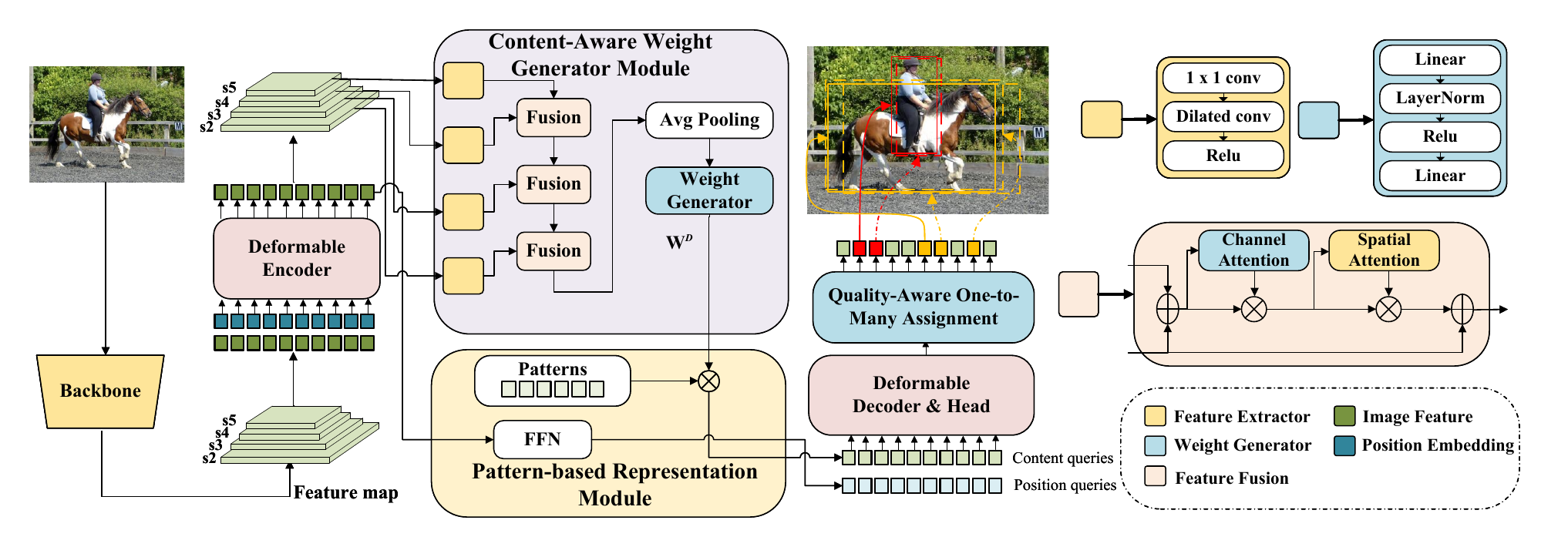}
\caption{
Overview of \textbf{PaQ-DETR}. 
Our framework integrates 
(1) a content-aware weight generator that adapts query composition to image features, 
(2) a pattern-based representation module that learns shared semantic bases, and 
(3) a quality-aware one-to-many assignment that provides balanced supervision.
}
\vspace{-1.0em} 
\label{fig:overview}
\end{figure*}

\noindent\textbf{DETR and Query Design.}
DETR~\cite{detr} formulates object detection as a set prediction task trained with one-to-one Hungarian matching, using a fixed set of learnable queries shared across all images.
Subsequent efforts improve convergence and adaptivity.
Deformable-DETR~\cite{deformable} introduces multi-scale deformable attention and generates encoder-derived proposals as content-dependent queries.
Conditional DETR~\cite{meng2021conditional} decomposes each query into content and spatial components, and Anchor DETR~\cite{wang2022anchor} injects spatial priors through anchor initialization.
Later works such as DINO~\cite{dino} observed that encoder-driven queries often suffer from semantic instability and reverted to purely learnable queries combined with denoising and contrastive objectives.
DN-DETR~\cite{li2022dn} and DAB-DETR~\cite{liu2022dab} further stabilize optimization via denoising and anchor parameterization.
RT-DETR~\cite{lv2023detrs} and RT-DETRv2~\cite{lv2024rt} reintroduce content-dependence by selecting top-$K$ encoder tokens as decoder queries for efficient adaptivity.
Overall, static queries provide stability but limited flexibility, whereas content-dependent queries improve adaptivity at the cost of semantic consistency.
PaQ-DETR bridges these paradigms by combining the semantic stability of static queries with the adaptivity of content-dependent queries.

\noindent\textbf{Dynamic and Pattern-Based Queries.}
DETR models often suffer from uneven query activation, where only a small subset of queries actively contribute to detection while remainings redundant~\cite{zhang2023dense,gao2024ease,chen2023enhanced,huang2024deim}.
This imbalance motivates exploration of dynamic mechanisms.
Dynamic DETR~\cite{dai2021dynamic} adaptively aggregates features at the head level.
DDQ-DETR~\cite{zhang2023dense} constructs queries as static combinations of shared bases, encouraging diversity yet remaining image-independent.
Dy-DETR~\cite{cui2023learning} fuses 900 queries into 300 using encoder-conditioned fusion to reduce redundancy, while EASE-DETR~\cite{gao2024ease} mitigates query competition via attention routing.
DQ-DETR~\cite{huang2024dq} dynamically adjusts the number of queries according to object density to improve small-object recall.
Although these works enhance efficiency and diversity, they often overlook semantic coherence between dynamic and shared queries.
Pattern-based formulations learn shared representations as reusable bases~\cite{wang2021a,wang2004face}.
Our approach extends it with content-dependent weighting to achieve both semantic stability and per-image adaptivity.

\noindent\textbf{One-to-Many and Hybrid Assignments.}
Beyond query formulation, several DETR variants revisit label assignment to enrich the sparse supervision of one-to-one matching.
Group-DETR~\cite{chen2023group} divides queries into fixed groups per ground truth,
MS-DETR~\cite{msdetr} applies mixed supervision across decoder stages,
and H-DETR~\cite{hdetr} introduces hybrid auxiliary branches to increase positive samples.
Co-DETR~\cite{zong2023codetr} proposes collaborative hybrid assignments that jointly optimize one-to-one and one-to-many branches through gradient sharing.
While these strategies accelerate convergence, they rely on fixed group sizes or additional decoders, and lack adaptive control over supervision quality.
We integrate pattern-based dynamic query generation with a quality-aware adaptive assignment, jointly addressing query and supervision imbalance in DETR training.

\section{Methodology}
\label{methodology}

PaQ-DETR is designed from a unified optimization perspective to alleviate the structural \emph{query activation imbalance} in one-to-one matching. This imbalance causes only a few “winning’’ queries to dominate gradient updates, leaving most queries weakly optimized. Alleviating this issue requires addressing two complementary aspects of DETR training: 
(1) the \emph{representation} of object queries, which determines how gradients propagate across queries, and 
(2) the \emph{distribution of supervision}, which determines how many queries receive meaningful training signals.

Accordingly, PaQ-DETR introduces two synergistic components. 
First, a \textbf{pattern-based dynamic query module} constructs each query as a convex combination of shared patterns using encoder-conditioned weights, enabling gradient sharing and stabilizing query semantics. 
Second, a \textbf{quality-aware adaptive assignment mechanism} adjusts both the number and selection of positive samples according to prediction quality, enriching supervision without auxiliary decoders. 
Together, these components form a coherent and lightweight framework that jointly improves query adaptivity and supervision balance.
We next describe the preliminaries and each component in detail.

\subsection{Preliminary}
\label{sec:preliminary}

\noindent\textbf{DETR Architecture.}
A DETR-style detector typically consists of three components: a convolutional backbone, a transformer encoder–decoder, and task-specific prediction heads.
Given an input image $\mathbf{I} \in \mathbb{R}^{h \times w \times 3}$, the backbone extracts image tokens $\mathbf{X} \in \mathbb{R}^{m \times d}$, which are refined by the encoder via self-attention to produce contextualized features $\mathbf{Z} = \text{Encoder}(\mathbf{X}) \in \mathbb{R}^{m \times d}$.
The decoder operates on a set of object queries $\mathbf{Q} = \{q_1, q_2, \dots, q_n\} \in \mathbb{R}^{n \times d}$, each of which iteratively attends to image features through self- and cross-attention across $L$ layers:
\begin{equation}
\mathbf{Q}^l = \text{FFN}(\text{CrossAttn}(\text{SelfAttn}(\mathbf{Q}^{l-1}), \mathbf{Z})).
\label{eq:inference1}
\end{equation}
The final layer output $\mathbf{Q}^L$ is mapped by prediction heads to class probabilities and bounding boxes $\mathbf{Y} = \text{Head}(\mathbf{Q}^L)$.
This formulation eliminates non-maximum suppression and allows end-to-end optimization through bipartite matching, which forms the basis for our dynamic query design.

\noindent\textbf{One-to-Many Assignment.}
DETR formulates object detection as a bipartite matching problem, where each ground-truth object is uniquely paired with a single prediction. The optimal matching $\sigma$ is obtained by minimizing a cost that jointly considers classification and bounding-box regression through Hungarian algorithm~\cite{kuhn1955hungarian}.
The one-to-one matching loss is expressed as:
\begin{equation}
\mathcal{L}_{1:1} =
\lambda_1 \mathcal{L}_{cls}(c_{\sigma}, c^*)
+ \lambda_2 \mathcal{L}_{box}(b_{\sigma}, b^*)
+ \lambda_3 \mathcal{L}_{giou}(b_{\sigma}, b^*),
\label{eq:11loss}
\end{equation}
where $\mathcal{L}_{cls}$ denotes the classification loss between the matched prediction $c_{\sigma}$ and ground-truth label $c^*$, and $\mathcal{L}_{box}$ and $\mathcal{L}_{giou}$ are the $L_1$ and GIoU losses between the matched boxes $b_{\sigma}$ and $b^*$, respectively. 

Although one-to-one strategy enables clean end-to-end optimization, it provides extremely sparse supervision—each ground truth updates only a single query. To enrich training signals, recent works extend to one-to-many assignment~\cite{cai2023align,msdetr}, allowing multiple predictions to match the same object.
The loss is accordingly reformulated as:
\begin{equation}
\begin{split}
\mathcal{L}_{1:m} =
& \sum_{i=1}^{k} \big[
\lambda_1 \mathcal{L}_{cls}(c_{\sigma(i)}, c^*)
+ \lambda_2 \mathcal{L}_{box}(b_{\sigma(i)}, b^*) \\
& \hspace{2em} + \lambda_3 \mathcal{L}_{giou}(b_{\sigma(i)}, b^*)
\big],
\end{split}
\label{eq:1mloss}
\end{equation}
where $\sigma(i)$ denotes the index of the $i$-th matched prediction among $k$ positive samples for the same ground-truth pair $(c^*, b^*)$. This relaxation improves convergence and generalization, but still relies on fixed number $k$ of positives.

\subsection{Dynamic Query Learning via Shared Patterns}
\label{subsec:subspacequery}

As illustrated in Fig.~\ref{fig:overview}, our framework extends DETR with a dynamic query generation mechanism that constructs image-specific queries from a compact set of shared base patterns.
The overall process consists of two components on the representation side: a content-aware weight generator that produces adaptive combination weights from encoder features, and a pattern-based module that learns reusable semantic bases for constructing object queries.
Together, these components enable queries to adapt to image content while preserving global semantic consistency and promoting balanced optimization.

\noindent\textbf{Query Representation via Shared Patterns.}
We represent object queries using a dual formulation comprising content queries $\mathbf{Q}^C = \{q_1^C, \dots, q_n^C\} \in \mathbb{R}^{n \times d}$ and base patterns $\mathbf{Q}^P = \{q_1^P, \dots, q_m^P\} \in \mathbb{R}^{m \times d}$, where $n$ and $m$ denote their respective quantities and $d$ is the embedding dimension.
Base patterns act as shared semantic primitives, while each content query $q_i^C$ is expressed as a convex combination of these patterns:
\begin{equation}
q_i^C = \sum_{j=1}^{m} w_{ij}^D q_{j}^{P},
\label{eq:query_combination}
\end{equation}
where $\mathbf{W}^D \in \mathbb{R}^{n \times m}$ contains the dynamic combination weights. Each $w_{ij}^D$ reflects the contribution of pattern $q_j^P$ to query $q_i^C$, constrained by $w_{ij}^D \ge 0$ and $\sum_{j=1}^{m} w_{ij}^D = 1$ to ensure a valid convex mixture. 
The learning of object queries is thereby transferred to the learning of shared base patterns. This formulation allows gradients from matched queries to flow through shared patterns, thereby promoting parameter sharing and alleviating the winner-take-all dynamic observed in independently optimized queries.
This pattern-based construction thus mitigates the representation-side imbalance by enabling more uniform gradient propagation across queries.

\noindent\textbf{Content-Aware Weight Generator.}
The content-aware weight generator module produces dynamic weights $\mathbf{W}^D$ by processing multi-scale encoder features through three lightweight components: 
feature extraction, multi-scale feature fusion, and weight generation. 
This module enables the learned patterns to be adaptively combined according to the input image.

\textit{Feature extraction.}
Given unflattened multi-scale feature maps $\mathbf{S}_i \in \mathbb{R}^{h_i \times w_i \times d}$ 
from the encoder outputs, 
we first apply a $1\times1$ convolution followed by a dilated convolution with ReLU activation to enlarge the receptive field and capture long-range dependencies that are essential for multi-object detection. This step enhances the non-linear ability of encoder features while maintaining efficiency.  

\textit{Multi-scale feature fusion.}
We perform top-down fusion by upsampling higher-level (lower-resolution) features to match the spatial resolution of lower-level (higher-resolution) ones, and then combine them through element-wise addition. 
The fused features are refined using lightweight channel attention~\cite{wang2020eca} and spatial attention~\cite{hou2021coordinate} to emphasize informative regions, while skip connections preserve low-level feature details. 
This process is applied progressively from top to bottom, producing an enhanced feature representation $\mathbf{Z} \in \mathbb{R}^{h_2 \times w_2 \times d}$ that encodes comprehensive multi-scale contextual information.  

\textit{Weight generation.}
The enhanced features $\mathbf{Z}$ are spatially aggregated using average pooling to obtain compact feature representations $\hat{\mathbf{Z}} \in \mathbb{R}^d$. 
A two-layer MLP $F_w$ with layer normalization and ReLU activation then transforms these features into dynamic weights:
\begin{equation}
\mathbf{W}^D = \text{softmax}(F_w(\hat{\mathbf{Z}})).
\end{equation}
The softmax operation ensures that each row of $\mathbf{W}^D$ sums to one, forming valid convex combinations that adaptively blend base patterns from the input image. 

\subsection{Quality-Aware One-to-Many Assignment}
\label{subsec:qualawareassign}
While the pattern-based module alleviates representation-side imbalance, the supervision-side imbalance caused by one-to-one matching remains a limiting factor for efficient optimization. Conventional DETR-based models~\cite{detr} adopt a one-to-one Hungarian matching scheme, where each ground-truth object supervises only a single prediction. While this formulation ensures end-to-end bipartite matching, it inherently limits the number of positive samples per object, leading to sparse supervision and slower convergence.
Recent variants~\cite{cai2023align, chen2023group, hdetr, msdetr} extend this paradigm to one-to-many matching to provide richer learning signals. Building on these insights, we propose a quality-aware one-to-many assignment strategy that dynamically determines both the number and selection of positive samples for each ground-truth instance based on prediction quality.

\noindent\textbf{Dynamic Positive Sample Selection.}
Given a set of predictions $\mathbf{\hat{P}} = \{\hat{p}_1, \hat{p}_2, \ldots, \hat{p}_n\}$, where each $\hat{p}_i = (\hat{b}_i, \hat{c}_i)$ represents a predicted bounding box and its classification confidence, and ground-truth objects $\mathbf{G} = \{g_1, g_2, \ldots, g_m\}$, we define a quality score for each prediction–ground-truth pair:
\begin{equation}
s_{i,j} = \text{IoU}(\hat{b}_i, g_j) - \gamma \hat{c}_i,
\end{equation}
where $\gamma$ controls the trade-off between localization accuracy and classification confidence. Higher $s_{i,j}$ values indicate higher-quality predictions.

For each ground truth $g_j$, the number of assigned positive predictions $k_j$ is adaptively determined as:
\begin{equation}
k_j = \max\!\left(\!\left\lceil \sum_{i \in \text{top-k}(s_{\cdot,j})} s_{i,j} \right\rceil,\, l\!\right),
\label{eq:quality_aware_k}
\end{equation}
where $\text{top-k}(s_{\cdot,j})$ indexes the top-$k$ predictions with the highest quality scores, and $l$ denotes the minimum number of positives per instance. This formulation allows the supervision strength to adapt to the confidence–localization consistency of predictions.

\noindent\textbf{Training Strategy.}
Our quality-aware assignment naturally prioritizes high-IoU but under-confident predictions, encouraging the model to focus on informative yet challenging samples. 
We apply this dynamic assignment in the intermediate decoder layers to enhance optimization robustness, while preserving standard one-to-one matching in the final layer for efficient inference.
For loss computation, we adopt the IoU-aware Varifocal Loss~\cite{zhang2021varifocalnet}, which adaptively weights positive samples by the quality scores, providing smoother gradients and more balanced supervision across varying sample difficulties.

\subsection{Overall Loss Function}
Our training objective consists of three parts: a quality-aware one-to-many assignment loss, a diversity regularization loss over shared patterns, and standard auxiliary Hungarian losses.

\noindent\textbf{Quality-Aware Assignment Loss.}
We adopt the proposed quality-aware one-to-many matching to construct adaptive positive sets. The loss $\mathcal{L}_{1:m}$ follows Eq.~\ref{eq:1mloss} with the coefficient settings from DINO~\cite{dino}.

\noindent\textbf{Pattern Diversity Regularization Loss.}
To avoid redundant pattern representations, we penalize cosine similarity among normalized shared patterns:
\begin{equation}
\mathcal{L}_{div} = 
\frac{1}{m(m-1)} \sum_{i \neq j} 
\big| \cos(\hat{q}_i^P, \hat{q}_j^P) \big|.
\end{equation}

\noindent\textbf{Auxiliary Supervision Loss.}
Following DETR, we apply Hungarian losses $\mathcal{L}_{aux}$ to all decoder intermediate layers.

\noindent\textbf{Total Objective Loss.}
The final loss is:
\begin{equation}
\mathcal{L}_{total} = 
\mathcal{L}_{1:m} + \mathcal{L}_{aux} + \beta \mathcal{L}_{div}.
\end{equation}

\section{Experiments}
\label{sec:experiments}

\begin{table*}[t]
\centering
\footnotesize
\caption{The performance on COCO \texttt{val}2017 based on ResNet-50 backbone under 12 and 24 epochs. ++ denotes we re-implement the methods and report the corresponding results. Rows marked in gray show the results integrating our methods.}
\resizebox{\linewidth}{!}{
\begin{tabular}{l|c|cc|cccccc}
\toprule
Method & Backbone & Epochs & Queries & \textrm{mAP} & $\textrm{AP}_{50}$ & $\textrm{AP}_{75}$ & $\textrm{AP}_S$ & $\textrm{AP}_M$ & $\textrm{AP}_L$ \\ 
\midrule
H-DETR~\cite{hdetr} & ResNet-50 & 12 & 300 & 48.7 & 66.4 & 52.9 & 31.2 & 51.5 & 63.5 \\
MS-DETR~\cite{msdetr} & ResNet-50 & 12 & 300 & 48.8 & 66.2 & 53.2 & 31.5 & 52.3 & 63.7 \\
Deformable-DETR++~\cite{deformable} & ResNet-50 & 12 & 300 & 46.9 & 65.6 & 51.1 & 30.1 & 50.4 & 60.3 \\
\rowcolor{gray!25}PaQ-Deformable-DETR (ours) & ResNet-50 & 12 & 300 & 48.4 & 66.9 & 52.6 & 32.2 & 52.2 & 63.1 \\
DAB-DETR++~\cite{liu2022dab} & ResNet-50  & 12 & 300 & 48.0 & 66.2 & 52.4 & 31.9 & 51.4 & 61.7 \\
\rowcolor{gray!25}PaQ-DAB-DETR (ours) & ResNet-50 & 12 & 300 & \textbf{49.2} & \textbf{67.7} & \textbf{53.6} & \textbf{32.8} & \textbf{52.6} & \textbf{63.9} \\
DN-DETR++~\cite{li2022dn} & ResNet-50 & 12 & 300 & 47.3 & 65.1 & 51.5 & 29.7 & 51.1 & 62.1 \\
\rowcolor{gray!25}PaQ-DN-DETR (ours) & ResNet-50 & 12 & 300 & 48.9 & 66.8 & 53.2 & 32.6 & 52.2 & 62.4 \\

\midrule
Cascade-DETR~\cite{ye2023cascade} & ResNet-50 & 12 & 900 & 49.7 & 67.1 & 54.1 & 32.4 & 53.5 & 65.1 \\
Stable-DINO~\cite{liu2023detection} & ResNet-50 & 12 & 900 & 50.4 & 67.4 & 55.0 & 32.9 & 54.0 & 65.5 \\
Align-DETR~\cite{cai2023align} & ResNet-50 & 12 & 900 & 50.2 & 67.8 & 54.4 & 32.9 & 53.3 & 65.0 \\
DDQ-DETR~\cite{zhang2023dense} & ResNet-50  & 12 & 900 & 50.7 & 68.1 & 55.7 & - & - & - \\
Salience-DETR~\cite{hou2024salience} & ResNet-50 & 12 & 900 & 49.2 & 67.1 & 53.8 & 32.7 & 53.0 & 63.1 \\
Rank-DETR~\cite{pu2024rank} & ResNet-50 & 12 & 900 & 50.4 & 67.9 & 55.2 & 33.6 & 53.8 & 64.2 \\
DAC-DETR~\cite{hu2024dac} & ResNet-50 & 12 & 900 & 50.0 & 67.6 & 54.7 & 32.9 & 53.1 & 64.4 \\
Dy-DETR~\cite{cui2023learning} & ResNet-50 & 12 & 900 & 49.7 & 68.1 & 54.2 & - & - & - \\
Ease-DETR~\cite{gao2024ease} & ResNet-50 & 12 & 900 & 50.8 & 68.9 & 55.3 & 34.1 & 54.2 & 65.1 \\
Deformable-DETR++~\cite{deformable} & ResNet-50 & 12 & 900 & 47.6 & 65.8 & 52.1 & 32.2 & 51.1 & 61.4 \\
\rowcolor{gray!25}PaQ-Deformable-DETR (ours) & ResNet-50 & 12 & 900 & 48.9 & 67.1 & 53.3 & 33.2 & 52.6 & 63.6 \\
DAB-DETR++~\cite{liu2022dab} & ResNet-50 & 12 & 900 & 48.3 & 66.4 & 52.8 & 32.1 & 51.6 & 62.0 \\
\rowcolor{gray!25}PaQ-DAB-DETR (ours) & ResNet-50 & 12 & 900 & 49.4 & 67.7 & 53.7 & 33.4 & 52.8 & 64.0 \\
DN-DETR++~\cite{li2022dn} & ResNet-50 & 12 & 900 & 48.7 & 66.4 & 52.9 & 32.1 & 52.1 & 63.7 \\
\rowcolor{gray!25}PaQ-DN-DETR (ours) & ResNet-50 & 12 & 900 & 49.8 & 67.6 & 53.5 & 33.2 & 52.7 & 64.8 \\
DINO++~\cite{dino} & ResNet-50 & 12 & 900 & 50.3 & 67.9 & 55.3 & 34.1 & 53.7 & 63.7 \\
\rowcolor{gray!25}PaQ-DINO (ours) & ResNet-50 & 12 & 900 & \textbf{51.9} & \textbf{69.1} & \textbf{56.3} & \textbf{35.1} & \textbf{56.0} & \textbf{66.6} \\

\midrule
DDQ-DETR~\cite{zhang2023dense} & ResNet-50 & 24 & 900 & 52.0 & 69.5 & \textbf{57.2} & 35.2 & 54.9 & 65.9 \\
Stable-DINO~\cite{liu2023detection} & ResNet-50 & 24 & 900 & 51.5 & 68.5 & 56.3 & 35.2 & 54.7 & 66.5 \\
Align-DETR~\cite{cai2023align} & ResNet-50 & 24 & 900 & 51.3 & 68.2 & 56.1 & 35.5 & 55.1 & 65.6 \\
MS-DETR~\cite{msdetr} & ResNet-50 & 24 & 900 & 51.7 & 68.7 & 56.5 & 34.0 & 55.4 & 65.5 \\
DAC-DETR~\cite{hu2024dac} & ResNet-50 & 24 & 900 & 51.2 & 68.9 & 56.0 & 34.0 & 54.6 & 65.4 \\
DINO++~\cite{dino} & ResNet-50 & 24 & 900 & 50.9 & 68.6 & 55.8 & 34.8 & 54.1 & 65.2 \\
\rowcolor{gray!25}PaQ-DINO (ours) & ResNet-50 & 24 & 900 & \textbf{52.6} & \textbf{69.7} & 56.9 & \textbf{35.7} & \textbf{56.4} & \textbf{67.0} \\
\bottomrule
\end{tabular}
}
\vspace{-1.0em}
\label{tab:coco-resnet}
\end{table*}

\begin{table*}[t]
\centering
\small
\caption{The performance on COCO \texttt{val}2017 based on Swin-L backbone under 12 epochs.}
\begin{tabular}{l|c|cc|ccccccc}
\toprule
Method & Backbone & Epochs & Queries & \textrm{mAP} & $\textrm{AP}_{50}$ & $\textrm{AP}_{75}$ & $\textrm{AP}_S$ & $\textrm{AP}_M$ & $\textrm{AP}_L$ \\ 
\midrule
$\mathcal{H}$-Def-DETR~\cite{hdetr} & Swin-L & 12 & 300 & 55.9 & 75.2 & 61.0 & 39.1 & 59.9 & 72.2 \\
DINO~\cite{dino} & Swin-L & 12 & 900 & 56.8 & 75.4 & 62.3 & 41.1 & 60.6 & 73.5 \\
Salience-DETR~\cite{hou2024salience} & Swin-L & 12 & 900 & 56.5 & 75.0 & 61.5 & 40.2 & 61.2 & 72.8 \\
Rank-DETR~\cite{pu2024rank} & Swin-L & 12 & 900 & 57.6 & 76.0 & \textbf{63.4} & \textbf{41.6} & 61.4 & 73.8 \\ 
DAC-DETR~\cite{hu2024dac} & Swin-L & 12 & 900 & 57.3 & 75.7 & 62.7 & 40.1 & 61.5 & 74.4 \\
Stable-DINO~\cite{liu2023detection} & Swin-L & 12 & 900 & 57.7 & 75.7 & \textbf{63.4} & 39.8 & 62.0 & \textbf{74.7} \\
Relation-DETR~\cite{hou2024relation} & Swin-L & 12 & 900 & \textbf{57.8} & 76.1 & 62.9 & 41.2 & 62.1 & 74.4 \\
\rowcolor{gray!20}PaQ-DINO (ours) & Swin-L & 12 & 900 & \textbf{57.8} & \textbf{76.2} & 62.6 & 41.4 & \textbf{62.2} & 74.2 \\
\bottomrule
\end{tabular}
\vspace{-1.0em}
\label{tab:coco-swin}
\end{table*}

\begin{table}[t]
\centering
\footnotesize
\caption{The performance on COCO \texttt{val}2017 with one-to-many assignment DETRs using ResNet-50 backbone under 12 epochs.}
\resizebox{\linewidth}{!}{
\begin{tabular}{l|ccccccc}
\toprule
Method & \textrm{mAP} & $\textrm{AP}_{50}$ & $\textrm{AP}_{75}$ & $\textrm{AP}_S$ & $\textrm{AP}_M$ & $\textrm{AP}_L$ \\ 
\midrule
Group-DETR~\cite{chen2023group} & 49.8 & - & - & 32.4 & 53.0 & 64.2 \\
$\mathcal{H}$-Def-DETR~\cite{hdetr} & 48.7 & 66.4 & 52.9 & 31.2 & 51.5 & 63.5 \\
MS-DETR~\cite{msdetr} & 50.0 & 67.3 & 54.4 & 31.6 & 53.2 & 64.0 \\
Co-DETR~\cite{zong2023codetr} & 52.1 & 69.4 & 57.1 & 35.4 & 55.4 & 65.9 \\
\rowcolor{gray!20}PaQ-DETR (ours) & \textbf{52.4} & \textbf{69.8} & \textbf{57.4} & \textbf{36.1} & \textbf{55.6} & \textbf{66.2} \\
\bottomrule
\end{tabular}
}
\label{tab:coco-1tom}
\end{table}

\subsection{Experiment Setup}
\noindent\textbf{Datasets.}
We perform comprehensive experiments to evaluate our model across multiple benchmarks and tasks, including the standard COCO 2017~\cite{coco}, CityScapes 2016~\cite{Cordts2016CityDataset}, CSD~\cite{wangcsd} and MSSD~\cite{ChenMSSD}. For object detection, we use COCO 2017 along with two task-specific datasets: CSD and MSSD. To further assess our approach's versatility, we extend our evaluation to instance segmentation using COCO 2017 and CityScapes 2016.

\noindent\textbf{Evaluation Metrics.}
Following established protocol~\cite{detr}, we evaluate detection performance using standard COCO metrics, including mean Average Precision (mAP) at different IoU thresholds (0.5, 0.75 and 0.5:0.95), as well as different scales across small, medium, and large objects. For instance segmentation, we report both mask mAP and box mAP metrics, respectively.

\noindent\textbf{Implementation details.}
We evaluate our approach using two backbone architectures: ResNet-50~\cite{he2016deep} pretrained on ImageNet-1k and Swin-Large~\cite{liu2021swin} pretrained on ImageNet-22k~\cite{deng2009imagenet}. All models are trained using the AdamW optimizer~\cite{loshchilov2017decoupled} with an initial learning rate of $2 \times e^{-4}$ and weight decay of $1\times e^{-4}$. We adopt standard training schedules of 1$\times$ (12 epochs) and 2$\times$ (24 epochs), where the learning rate is reduced by a factor of 0.1 at the 11th and 20th epochs, respectively. Training is performed with a batch size of 16 across 8 NVIDIA RTX 3090 GPUs. Following established practices~\cite{detr, deformable, dino}, we apply standard data augmentations, including random resize, crop, and horizontal flip during training. For dynamic query generation, we employ 50 and 150 patterns to generate 300 and 900 queries, respectively, consistent with baseline configurations. Within  one-to-many assignment,  the hyper-parameters are configured as follows: $l=1$, $k=4$, and $\gamma=0.4$. The diversity loss weight is set to $\beta=0.2$.

\begin{table}[t]
\centering
\footnotesize
\caption{The performance on CSD based on ResNet-50 backbone under 60 epochs.}
\begin{tabular}{l|ccccc}
\toprule
Method & \textrm{mAP} & $\textrm{AP}_{50}$ & $\textrm{AP}_{75}$ & $\textrm{AP}_S$ & $\textrm{AP}_M$ \\ 
\midrule
DAB-DETR~\cite{liu2022dab} & 50.2 & 88.5 & 51.0 & 48.2 & 38.2 \\
DN-DETR~\cite{li2022dn} & 49.9 & 88.0 & 51.2 & 47.6 & 37.7 \\
H-DETR~\cite{hdetr} & 53.0 & 90.6 & 55.7 & 51.2 & 39.2 \\
DINO~\cite{dino} & 53.4 & 91.0 & 55.8 & 50.9 & 39.6 \\
\rowcolor{gray!25}PaQ-DINO (ours) & \textbf{54.2} & \textbf{91.8} & \textbf{56.2} & \textbf{51.4} & \textbf{40.2} \\
\bottomrule
\end{tabular}
\label{tab:csd}
\end{table}

\begin{table}[t]
\centering
\footnotesize
\caption{The performance on MSSD based on ResNet-50 backbone under 120 epochs.}
\begin{tabular}{l|ccccc}
\toprule
Method & \textrm{mAP} & $\textrm{AP}_{50}$ & $\textrm{AP}_S$ & $\textrm{AP}_M$ & $\textrm{AP}_L$ \\ 
\midrule
DAB-DETR~\cite{liu2022dab} & 33.7 & 60.0 & 15.9 & 26.7 & 29.0 \\
DN-DETR~\cite{li2022dn} & 45.6 & 74.1 & 18.6 & 31.9 & 41.0 \\
H-DETR~\cite{hdetr} & 46.9 & 76.8 & 20.3 & 45.3 & 40.7 \\
DINO~\cite{dino} & 51.0 & 80.0 & 20.0 & 47.4 & 44.8 \\
\rowcolor{gray!25}PaQ-DINO (ours) & \textbf{55.2} & \textbf{80.2} & \textbf{27.2} & \textbf{48.2} & \textbf{45.2} \\
\bottomrule
\end{tabular}
\label{tab:mssd}
\end{table}

\subsection{Main Results}
\label{subsec:mainresults}

\noindent\textbf{COCO 2017.}
We evaluate our methods on the COCO 2017 benchmark~\cite{coco} across multiple DETR baselines, including Deformable-DETR~\cite{deformable}, DAB-DETR~\cite{liu2022dab}, DN-DETR~\cite{li2022dn}, and DINO~\cite{dino}, using 300 or 900 queries.
Unlike Co-DETR~\cite{zong2023codetr}, we follow the standard DETR training pipeline without introducing extra queries or auxiliary branches. 
As shown in Table~\ref{tab:coco-resnet}, our PaQ variants consistently improves all baselines by 1.1–1.6 mAP under the 12-epoch schedule. 
Notably, PaQ-DINO reaches 51.9 mAP, surpassing DINO++ by 1.6 and yielding larger gains on medium (+2.3 AP$_M$) and large (+2.9 AP$_L$) objects. 
With 24-epoch training, it further achieves 52.6 mAP, outperforming recent methods such as DDQ-DETR~\cite{zhang2023dense} and Stable-DINO~\cite{liu2023detection}. 
Using a Swin-L backbone (Table~\ref{tab:coco-swin}), PaQ-DINO attains 57.8 mAP, surpassing all counterparts.
These results demonstrate that our dynamic query and quality-aware designs generalize well across architectures and training schedules.

\noindent\textbf{Comparison with One-to-Many Assignment DETRs.}
Recent DETR variants~\cite{zong2023codetr,msdetr,chen2023group,hdetr} employ one-to-many assignment to enrich supervision and speed up convergence. 
To ensure a fair comparison, we follow the hybrid configuration of H-DETR~\cite{hdetr} only in training with $N_h\!=\!1500$ and $k\!=\!6$ positives in an auxiliary branch, while keeping our main branch equipped with the proposed adaptive assignment, termed as PaQ-DETR.
As shown in Table~\ref{tab:coco-1tom}, PaQ-DETR achieves the highest mAP among representative one-to-many methods, including Group-DETR~\cite{chen2023group}, MS-DETR~\cite{msdetr}, and Co-DETR~\cite{zong2023codetr}, confirming that our unified pattern- and quality-aware design provides complementary benefits for feature diversity and balanced supervision.

\noindent\textbf{CSD and MSSD.}
To demonstrate generalizability, we evaluate our method on two task-specific datasets for defect detection: CSD~\cite{wangcsd} and MSSD~\cite{ChenMSSD}. As shown in Table~\ref{tab:csd} and Table~\ref{tab:mssd}, our method achieves 54.2 and 55.2 mAP respectively, improving over DINO baseline by 0.8 and 4.2 mAP. These consistent gains across diverse datasets validate the robustness of our method beyond general object detection.

\subsection{Ablation Study}
\label{sec:ablationstudy}

\begin{figure}
(a)~~\includegraphics[width=0.17\textwidth,height=0.08\textheight]{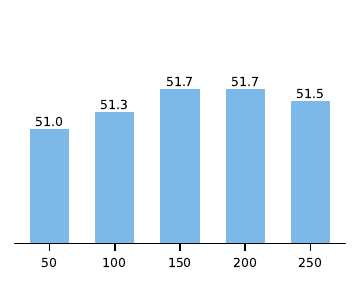}~~
(b)~~\includegraphics[width=0.17\textwidth,height=0.08\textheight]{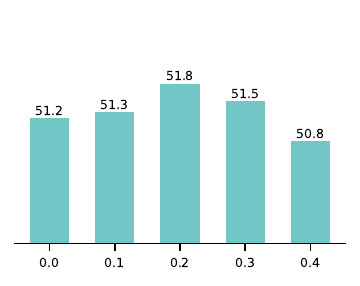}~~\\[-0.5em]
(c)~~\includegraphics[width=0.17\textwidth,height=0.08\textheight]{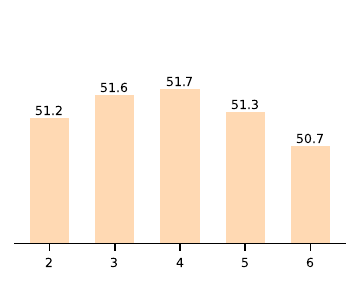}~~
(d)~~\includegraphics[width=0.17\textwidth,height=0.08\textheight]{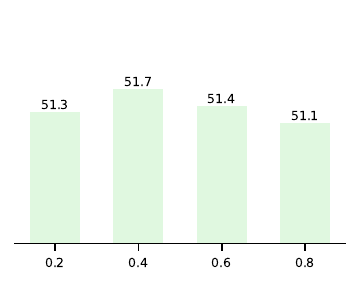}
\caption{
Ablation study on key hyperparameters of PaQ-DETR:
(a) number of patterns, (b) diversity loss weight $\beta$, 
(c) top-$k$ in quality-aware assignment, and (d) balancing weight $\gamma$.
}
\label{fig:ablation}
\end{figure}

\noindent\textbf{Pattern number analysis.}
Fig.~\ref{fig:ablation}(a) shows the optimal performance 51.7 mAP is achieved with 150 or 200 patterns. Even a compact set of 50 patterns still yields +0.7 mAP over the baseline, suggesting that a small number of shared patterns is sufficient to capture diverse object semantics.
However, 250 patterns cause slight degradation due to optimization interference from excessive pattern diversity.

\noindent\textbf{Diversity loss weight analysis.}
Fig.~\ref{fig:ablation}(b) illustrates that removing the diversity regularization ($\beta\!=\!0$) leads to a 0.6 mAP drop, while too strong a weight ($\beta\!>\!0.3$) also harms accuracy. 
A moderate $\beta\!=\!0.2$ achieves a good trade-off between stability and diversity.

\noindent\textbf{One-to-many assignment hyper-parameters.}
For the one-to-many assignment in Fig.~\ref{fig:ablation}(c–d), the best results occur at top-k selection $k\!=\!4$ and IoU-classification balance $\gamma\!=\!0.4$. Larger $k$ values introduce low-quality matches, confirming the benefit of positive assignments with selective, quality-driven supervision.

\begin{table}[t]
\centering
\footnotesize
\caption{Ablation study on different components. `D': dynamic query, `Q': Quality-aware one-to-many assignment.}
\resizebox{\linewidth}{!}{
\begin{tabular}{cc|ccccccc}
\toprule
D & Q & \textrm{mAP} & $\textrm{AP}_{50}$ & $\textrm{AP}_{75}$ & $\textrm{AP}_S$ & $\textrm{AP}_M$ & $\textrm{AP}_L$ & Gini \\ 
\midrule
& & 50.3 & 67.9 & 55.3 & 34.1 & 53.7 & 63.7 & 0.97 \\
\checkmark & & 51.4 & 69.0 & 56.0 & \textbf{35.3} & 55.2 & 66.5 & 0.90 \\
&  \checkmark & 51.1 & 68.4 & 55.5 & 34.8 & 55.1 & 65.6 & 0.95 \\
\checkmark &  \checkmark & \textbf{51.9} & \textbf{69.1} & \textbf{56.3} & 35.1 & \textbf{56.0} & \textbf{66.6} & 0.89 \\
\bottomrule
\end{tabular}
}
\label{tab:abl:component}
\end{table}

\noindent\textbf{Component ablation.}
Table~\ref{tab:abl:component} reports the effect of each component over the DINO++ baseline.
Dynamic query learning alone improves performance by +1.1 mAP with clear gains on large objects (+2.8 AP$_L$). Quality-aware assignment contributes an additional +0.8 mAP. When combined, our method reaches 51.9 mAP, notably boosting AP$_M$ (+2.3) and AP$_L$ (+2.9). These results confirm the complementary benefits of dynamic query generation and quality-aware supervision.
In addition to accuracy, we also observe a reduction in query imbalance, as reflected by the lower Gini coefficient (0.97 → 0.89), demonstrating that our design encourages more balanced query utilization during training.

\subsection{Model Analysis}
\begin{figure}[t]
\centering
\includegraphics[scale=0.5,trim= 0.5pt 0.5pt 0.5pt 1.0pt,clip]{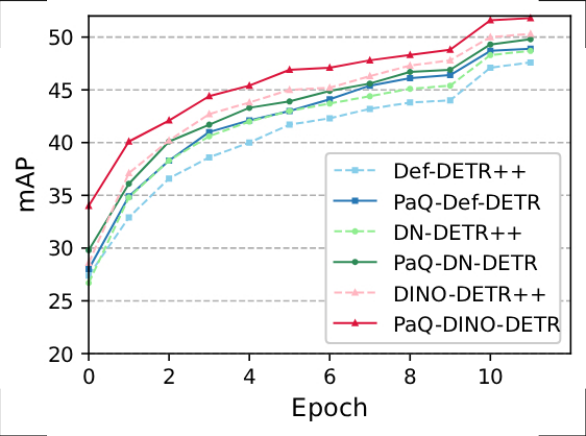}
\caption{Convergence curves comparing baseline (dashed lines) with our methods (solid lines). The horizontal axis shows the number of the epochs, while the vertical axis shows mAP metrics.}
\label{fig:convergence-curve}
\end{figure}

\noindent\textbf{Analysis of convergence.}
Fig.~\ref{fig:convergence-curve} compares convergence curves with a ResNet-50 backbone (900 queries, 12 epochs). PaQ variants converge faster and achieve higher accuracy than Def-DETR, DN-DETR, and DINO. The dynamic query learning provides stronger initialization and stabilizes optimization, explaining the observed acceleration over static-query counterparts.

\begin{figure}[t]
\centering
(a)\includegraphics[width=0.9\columnwidth]{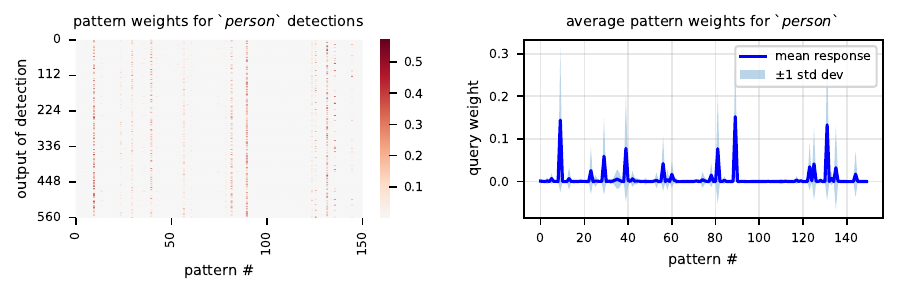}
(b)\includegraphics[width=0.9\columnwidth]{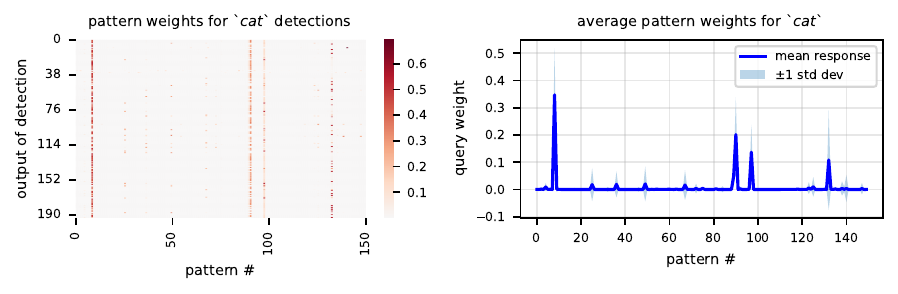}
\caption{Pattern activation visualization. Left: activation heat-map; Right: statistical distribution of weights over successful detection for (a) person and (b) cat categories.}
\label{fig:pattern}
\end{figure}
\begin{figure}[t]
\centering
\includegraphics[width=0.45\textwidth,height=0.35\textwidth]{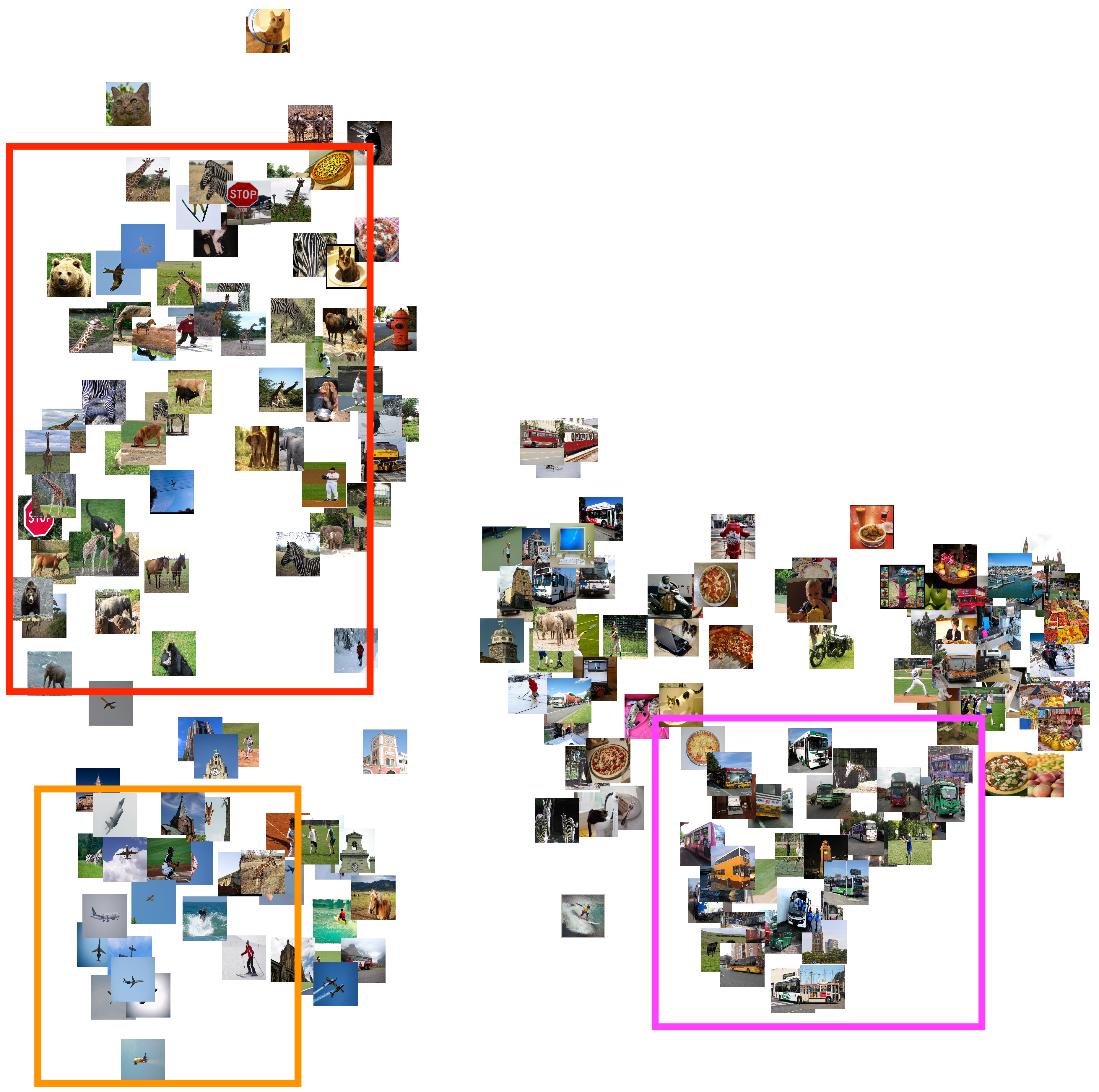}
\caption{t-SNE visualization of $\mathbf{W}^D$ on 200 images from COCO \texttt{val}2017. Zoom in to see details.}
\vspace{-1.0em}
\label{fig:tsne}
\end{figure}

\noindent\textbf{Analysis of pattern activations.}
We analyze the activation behavior of dynamic weights $\mathbf{W}^D$ that transform learned patterns into dynamic queries. As shown in Fig.~\ref{fig:pattern}, we examine two representative categories, \textit{person} and \textit{cat}, on COCO \texttt{val}2017, considering successful detections with $\text{IoU}>0.7$ and confidence $>0.5$. The activations are highly sparse, indicating that detection relies on selective pattern combinations. Interestingly, \textit{person} and \textit{cat} share partially overlapping activations, suggesting the presence of reusable patterns across categories.

\noindent\textbf{Analysis of semantic clustering.}
Because the dynamic weights $\mathbf{W}^D$ are conditioned on image features, semantically similar images should yield similar $\mathbf{W}^D$ distributions. We flatten $\mathbf{W}^D$ from 200 COCO validation images and project them using t-SNE for 2D visualization. As shown in Fig.~\ref{fig:tsne}, clear semantic clusters emerge—animals (red) occupy the top-left region, aircraft (amber) the bottom-left, and vehicles (purple) the bottom-right—indicating that PaQ-DETR encodes meaningful semantics when generating content-aware dynamic queries.

\subsection{Computational and Memory Efficiency}
We analyze the computational efficiency of PaQ-DETR in terms of parameters, FLOPs, memory, and inference speed. As shown in Table~\ref{tab:compute_resource}, our PaQ-DINO introduces marginal overhead compared to strong DETR baselines.
The additional cost is less than +5\% FLOPs and +0.5 GB memory, with only 0.2 FPS decrease in inference speed. It shows that our method preserves almost the same runtime while delivering consistent performance gains of +1.2–4.2 mAP across datasets.

\begin{table}[t]
\centering
\footnotesize
\caption{Computational efficiency analysis.}
\setlength{\tabcolsep}{5pt}
\resizebox{\linewidth}{!}{
\begin{tabular}{l|cccc}
\toprule
Method & Params (M) & FLOPs (G) & Mem (GB) & FPS \\
\midrule
Deformable-DETR++~\cite{deformable} & 40.5 & 188 & 3.63 & 16.4 \\
DN-DETR++~\cite{li2022dn} & 41.2 & 192 & 3.64 & 16.6 \\
DINO++~\cite{dino} & 42.0 & 196 & 3.67 & 15.8 \\
\rowcolor{gray!10}
\textbf{PaQ-DINO (ours)} & \textbf{43.8} & \textbf{205} & \textbf{4.15} & \textbf{15.4} \\
\bottomrule
\end{tabular}
}
\label{tab:compute_resource}
\end{table}

\subsection{Instance Segmentation}
We further extend our method to instance segmentation by incorporating an additional mask prediction head into the transformer-based architecture. We evaluate on COCO val2017~\cite{coco} and Cityscapes~\cite{Cordts2016CityDataset} using 300 queries, with Deformable-DETR~\cite{deformable} serving as baseline.
Table.~\ref{tab:insseg} presents both mask AP and box AP results under 12 and 24 epochs. On COCO \texttt{val}2017, we achieve substantial mask AP improvements of 2.4 (12 epochs) and 2.1 (24 epochs) over the baseline. Similarly, on Cityscapes, we observe consistent gains of 2.0 and 2.2 for mask AP at 12 and 24 epochs. The box AP results also show comparable improvements, indicating that our proposed method benefits both localization and segmentation.

\begin{table}[t]
\centering
\footnotesize
\caption{Instance segmentation results on COCO \texttt{val}2017 and CityScapes 2016 with 300 queries and ResNet-50 backbone.}
\begin{tabular}{l|c|c|c}
\toprule
Method & Epochs & Mask mAP & Box mAP \\
\midrule
\rowcolor{gray!25}\multicolumn{4}{l}{\texttt{\textbf{Dataset: COCO val2017}}} \\
\midrule
Baseline~\cite{deformable} & 12 & 32.4 & 46.5 \\
PaQ-Def-DETR (ours) & 12 & \textbf{34.8} & \textbf{48.4} \\
Baseline~\cite{deformable} & 24 & 35.1 & 48.6 \\
PaQ-Def-DETR (ours) & 24 & \textbf{37.2} & \textbf{50.6} \\
\midrule
\rowcolor{gray!25}\multicolumn{4}{l}{\texttt{\textbf{Dataset: CityScapes 2016}}} \\
\midrule
Baseline~\cite{deformable} & 12 & 34.8 & 52.6 \\
PaQ-Def-DETR (ours) & 12 & \textbf{36.8} & \textbf{54.8} \\
Baseline~\cite{deformable} & 24 & 36.6 & 54.4 \\
PaQ-Def-DETR (ours) & 24 & \textbf{38.8} & \textbf{56.2} \\
\bottomrule
\end{tabular}
\label{tab:insseg}
\end{table}

\section{Conclusion}

We introduced PaQ-DETR, a novel approach that enhances DETR-based object detection through two key innovations: dynamic query learning via shared patterns and quality-aware one-to-many assignment. Extensive experiments on COCO 2017, CityScapes 2016 and two specialized datasets, demonstrate consistent performance gains across various models and backbones. Our analysis reveals that object queries can be effectively represented by a smaller set of learnable patterns, and these patterns exhibit meaningful semantic clustering properties that adapt to image content.
{
    \small
    \bibliographystyle{ieeenat_fullname}
    \bibliography{main}
}


\end{document}